\documentclass{article} % For LaTeX2e
\usepackage{iclr2021_conference,times}
\usepackage{csquotes}

\usepackage{amsmath,amsfonts,bm}

\def\eqref#1{equation~\ref{#1}}
\def\1{\bm{1}}

\DeclareMathAlphabet{\mathsfit}{\encodingdefault}{\sfdefault}{m}{sl}
\SetMathAlphabet{\mathsfit}{bold}{\encodingdefault}{\sfdefault}{bx}{n}

\usepackage{natbib}
\usepackage{hyperref}
\usepackage{url}

\title{The Role of General Intelligence in \\ Mathematical Reasoning}

\author{Aviv Keren \thanks{\href{mailto:aviv.keren@gmail.com}{\texttt{aviv.keren@gmail.com}} %\url{https://avivkeren.wixsite.com/website}}}
\newline \hphantom \hspace{0.5cm} \url{https://avivkeren.wixsite.com/website}}}

\newenvironment{myquote}%
  {\list{}{\leftmargin=0.15in\rightmargin=0.15in}\item[]}%
  {\endlist}

\iclrfinalcopy % Uncomment for camera-ready version, but NOT for submission.
\begin{document}

\maketitle

\begin{abstract}
Objects are a centerpiece of the mathematical realm and our interaction with and reasoning about it, just as they are of the physical one (if not more). And humans' mathematical reasoning must ultimately be grounded in our general intelligence. Yet in contemporary cognitive science and A.I., the physical and mathematical domains are customarily explored separately, which allows for baking in assumptions for what objects \textit{are} for the system - and missing potential connections.

In this paper, I put the issue into its philosophical and cognitive context. I then describe an abstract theoretical framework for learning object representations, that makes room for mathematical objects \textit{on par} with non-mathematical ones. Finally, I describe a case study that builds on that view to show how our general ability for integrating different aspects of objects effects our conception of the natural numbers.

\end{abstract}

\section{Background – Mathematical Reasoning}
Getting A.I. to achieve some skill often forces us to gain a better, more concrete and detailed understanding of that skill and what it requires – and the skill of mathematical reasoning is no different. But what is the current state of our understanding of it?

There is no understanding of a form of reasoning without a decent understanding of the domain which that reasoning is about, but the case of mathematics makes it particularly challenging on the former. The common conception of mathematics, in philosophy of mathematics as well as amongst working mathematicians, is of a non-physical, non-psychological, timeless realm. This gives rise to the principled Benacerraf challenge: \textquote{[T]o provide an account of the mechanisms that explain how our beliefs about these remote entities can so well reflect the facts about them.}\citep{benacerraf1973mathematical,field1989realism}

A central component here would supposedly be mathematical language, its logic, to which us humans relate – however that language in turn relates to the mathematical realm. Yet limitations of (first-order) logic (both Gödelean incompleteness and the existence of non-standard models) prohibit it from grounding even a fundamental and seemingly simple structure such as $\mathbb{N}$. This nevertheless still standard (if orthodox)\footnote{A new wave of Martin-Löf type-theoretic approaches to the foundations of mathematics (esp. homotopy type theory) is on the rise \citep{hottbook,Grayson_2018}. While aiming to capture what mathematicians consciously intend, there is no presumption there to go \textit{beyond} what they can perceive, to reflect rich \textit{non-conscious} processing à la cognitive science and A.I.} logic-centered view also puts aside the question of how mathematics is applied, used.\footnote{To put these delicate matters extremely crudely. For general introductions, see the classic \citep{benacerraf_putnam_1984} or the modern \citep{shapirohandbook}}

This situation, however, does not deter mathematical cognition, the scientific field in charge of investigating and accounting for math-related reasoning, from exploring its topic. This field has been growing immensely in the past few decades.\footnote{For a modern intro, see e.g. \citep{intromathcog}} It reveals aspects of what mathematical objects are to us, that we may not be aware of. One such example is the distance effect, whereby the discrimination of numbers turns out to depend (logarithmically) on their magnitudes rather than their representation in digits. (e.g., it is in a sense harder to tell that 69\textless71 than that 61\textless69, in spite of the leading digit). More generally, we seem to possess an innate Approximate Number System, which is considered to underlie our so-called “number sense” and even our conception of number. This is in contrast to the aforementioned logical point of view, by which approximations seem irrelevant to the \textit{constitution} of numbers.

The contrast with a logic-based approach runs deep, for a few important reasons. With a substantial attention to not only human-general mathematical abilities, but those of infants and even many animals, the place of language seems more of a late addendum than any sort of substantial grounding of the mathematics involved. And the evolutionary pragmatic view downplays the absolute, precise nature of mathematics anyhow (the main ingredient for which logic seems to be the natural, even necessary, candidate). Instead, it accentuates mathematics’ “applied”, real life manifestation. And even for us adult humans, language is after all ordinarily only a referential expression of an independently existing, independently perceived domain. In particular here, we don’t seem to come to be acquainted with the natural numbers via some Peano Arithmetic like axiomatization (e.g. Two numbers of which the successors are equal are themselves equal), qua fact registration. In fact, we come to see such axioms \textit{as} true (and there is no obvious more-fundamental, psychologically-viable candidate axiomatization). Yet that we do grasp that number structure, sees like an undisputable starting point. As \cite{dehaene2011number} puts it:
%\begin{displayquote}[\citeyear{dehaene2011number}]
%\begin{quotation}
\begin{myquote}
Ironically, any 5-year-old has an intimate understanding of those very numbers that the brightest logicians struggle to define. No need for a formal definition: We know intuitively what integers are. Among the infinite number of models that satisfy Peano’s axioms, we can immediately distinguish genuine integers from other meaningless and artificial fantasies. Hence our brain does not rely on axioms.
\end{myquote}
%\end{quotation}
%\end{displayquote}
Dehaene attributes our basic numerical abilities and concepts to an \textit{innate} “number sense”. Against this common position, others take a more \textit{learnt} view. \citet{carey2009origin}, for example, proposes a theory of how we come to acquire concepts that ascend beyond a current conceptual capacity (numbers being a prime example), through a process of “Quinean bootstrapping” (in which symbols are initially only partly interpreted, while we nevertheless learn relations between them). Other positions in-between complete innateness and cultural acquisition abound (see e.g. \citep{spelkecore2017}), and the scientific issue is not settled \citep{Nunez2017}.

Computational models (including neural ones) for the distance effect and other findings can and are being given (e.g. \citep{computationaldistanceeffect}). But the current methodological limitations of cognitive science as an empirical field do not allow for approaching a sufficiently-definite account how of the related cognitive machinery actually functions, computationally. And so, the tension, between a mechanistically-grounded conception of numbers that somehow gives rise to their truths too, and the logical impossibility results (which apply quite generally), can be kept repressed, for now. %better connect
What in particular is still missing is how numbers, and mathematical objects in general, are represented (if not merely as the logic that’s about them, their theories).

\section{Learning General Object Representation}

Objects seem to take a central place in the structure of our mental realm, at least at its highest, conscious levels. Accordingly, how our representations of objects are \textit{learnt}, or what aspects of it are, rather, supported innately, is an important if not central topic and general theme in cognitive science. It is less eminent in contemporary A.I., although the field of developmental robotics, for example, \textquote{aims to show how a robot can start with the ‘blooming, buzzing confusion’ of low-level sensorimotor interaction, and can learn higher-level symbolic structures}\citep{modayil2008}; and although there is interest in locating the issue within contemporary settings \citep{neuripsobjrep}.

\textit{Mathematical} objects are different from other sorts of objects in fundamental respects. But we do consider them, and seem to come to master their grasp and manipulation, as objects. (Some even talk in \textit{literal} terms of perception \citep{godelcantor}.)
Ruling out a specific evolutionary adaptation suggests that this is done very much through pre-existing, general cognitive mechanisms. This poses, for both cognitive science and A.I., the central challenge that I wish to put forth here:

{\em To account for the learning of object representations in sufficiently abstract theoretical terms, so as to pertain over both mathematical and non-mathematical objects} (before getting into the differences, within such a system).

Such abstraction may be possible, I suggest, from an approach such as \citep{modayil2008}'s:
%\begin{displayquote}
%\begin{quotation}
\begin{myquote}
{[i]nstead of considering an object to be a physical entity, we consider an object to be an explanation for some subset of an agent’s experience [Or: “a hypothesized entity that accounts for a spatiotemporally coherent cluster of sensory experience.”]. With this approach, the semantics of an object are intrinsically defined from the agent’s sensorimotor experience"}
\end{myquote}
%\end{quotation}
%\end{displayquote}
While the interest in robotics is in the physical domain, of sensory experience and motoric influence within the spatiotemporal, the underlying notions and principles at play need not be limited to that. Experience, in particular, also includes experience with the internal workings and patterns within the cognitive system itself (to some extent); and coherence is certainly an abstract notion to begin with. In this respect, consider our related abilities with physical objects:
%\begin{displayquote}[\citep{kinzler2007core}]
%\begin{quotation}
\begin{myquote}
The core system of object representation centers on a set of principles governing object motion: cohesion (objects move as connected and bounded wholes), continuity (objects move on connected, unobstructed paths), and contact (objects influence each others’ motion when and only when they touch). These principles allow infants.. to perceive the boundaries and shapes of objects that are visible or partly out of view, and to predict when objects will move and where they will come to rest… research with older infants suggests that a single system underlies infants’ object representations. \citep{kinzler2007core}
\end{myquote}
%\end{quotation}
%\end{displayquote}
Even if such principles develop innately, they need not be coded in in such explicitly-physical terms. The theoretical challenge, then, is to subsume such principles under a general computational or information-theoretic notion. %of \textit{object-coherence}?

In \citep[Chapter~6]{AvivKerenPhD}, I develop a preliminary theoretical account of such an abstract framework (restricted to a specific aspect), describing the following:
\begin{enumerate}
\item The fundamentals regarding such a system’s treatment of types of objects (e.g. apples).
\item How such types can (in a technical sense) be extended, e.g. oranges as a kind of fruit that can be squeezed. Or combined, "amalgamated" with other types, e.g. bats as a kind of winged mammals; or smartphones as phones + computers + cameras + g.p.s...; or a 3-d structure perceived through two 2-d channels (eyes).
\item How these issues depend on the environment, on the regularities it presents. e.g., cast shadows that tend to appear under balls, being integrated into their very perception, that of their location in 3d-space.
\item What principles guide actual development – learning, mastering, and maintaining this sort of representation.
\end{enumerate}

The immense scope of a full such theory aside, my own restricted focus sets to substantiate the following picture: sufficiently regular interactions between represented, coherent object-types (e.g. a person's looks and their voice%do better
) allow for amalgamating them into a new coherent object-type, assimilating the interaction between the original object-types, that are re-conceived a\ \textit{aspects} of the new, "higher level" object-type. Such higher level object-types are more valuable, involving a dimensionality reduction and offering in particular an enlarged space of affordances. Thus, the system is directed towards pursuing such amalgamations whenever it detects empirical regularities that allow for it, structuring the perceived ontology accordingly.

On a methodological note: Beyond having to stand on its own merits as a reasonable 'story' of how such a system would operate, the full challenge for such a theoretical account is two-fold. On the one hand, it should be able to serve as the engineering starting-point for the construction of a working A.I. system (also showing how related topics such as binding \citep{greff2020binding} and disentanglement \citep{higgins2018definition} fit in). On the other, it should square empirically, as a generalization, with the many related issues from cognitive science (the core system of object representation, multi-modal integration, various forms of binding, categorization, and more); and hopefully produce new predictions as well.

A structurally related theory in cognitive science is that of \textit{conceptual blending} \citep{conceptual_blending}. It accounts for concepts such as \textit{houseboat} (as well as more temporary or contextual blends) as the blended "mental space" that shares some counterparts of the input spaces of e.g. house \& boat, aligned across a more abstract generic space that they both share. It offers optimality principles that govern the blending process. Blending has in particular been used to account for concepts in mathematics as well \citep{LakoffNunez00,GUHE2011249}.
Various computational implementations have followed the cognitive theory, formalizing its notions and principles \citep{EPPE2018105}; and even a theoretical Category Theoretic formulation that unifies these various A.I. concretizations \citep{SCHORLEMMER2021118}.\\
The phenomenon explored here, however, is treated as a very high-level (and rather peripheral) conceptual one, concerning (perhaps human-only) creativity. And the computational implementations are grounded in logic, accordingly. The different focus I suggest here is on such combination as a fundamental component of the cognitive construction of object-representations at large. In particular, how the regularities that govern the possibility and value of such combinations are handled, should be understood as part of the more general working of the system, picking up on and integrating \textit{statistical} patterns into the construction of its ontology. In terms of A.I., this reflect the on-going quest to find the proper place for the symbolic within modern architectures, which are chiefly neural.
%extra degrees of freedom, not restricted or directed by reality
%one striking example of the generality ..I suggest / is Dal-E, which is purely neural, yet seems to achieve a similar feat, perhaps even creatively-so. / seems to go beyond the limited semantic richness of the linguistic concepts themselves

Given the suggested picture, of how an object-centered system might in general come to attribute objecthood or break it apart, we can approach the development of our conception of numbers and account for a hidden, sub-symbolic intricacy in that conception.

\section{Case Study: Ordinals and Cardinals in the Natural Numbers} %$\mathbb{N}$ 
The common mathematical narrative is that only in going to infinity, numbers had to broken up into ordinals and cardinals (as representatives of order-type and of quantity). In the finite, these are customarily considered as merely different aspects of the very same \textit{natural} numbers, in how they are used. This narrative can be argued to be conceptually problematic, however, given how distinct ordinals and cardinals turned out to be, not only conceptually and in terms of their usage, but ontologically and arithmetically.

\citet{steiner2005applicability} substantiates this doubt with an analysis of the intricate interplay between ordinals and cardinals already \textit{in the finite realm}. We learn to transition between these "aspects" from the very beginning, where counting (through some particular ordering) is to produce the (order-independent) quantity (the so-called "Cardinality Principle"\citep{cardinalprinciple,cp-knower}). But the child may still wonder, \textquote{Why is it always the case that adding x, y times, gives the same result as adding y, x times?} (that is, in terms of orderings, with addition as concatenation, not quantities and disjoint union). And she would be "right" to wonder, since the commutativity of multiplication is indeed \textbf{not} a property of ordinal arithmetic, only cardinal arithmetic (if evident only in the infinite). Steiner gives a multi-stage analysis of how we transition from ordinals to cardinals, commute there through the (cardinality) isomorphism between the Cartesian products $X\times Y$ and $Y\times X$, and return to ordinals, in order to see that fact.

Steiner's analysis reveals finer workings within a seemingly extremely familiar and basic realm. Though his account does not concern with cognition or A.I., we can take it to show that we were consciously missing this complexity that our brains have nonetheless learned to manage successfully and systematically. In \citep[Chapter~7]{AvivKerenPhD}, I account for this in terms of the framework described in the previous section, as follows.

If ordinals and cardinals be amalgamated ("combined") into numbers, then by the time we acquire the latter, the former object-types must conceivably already be grasped by the system (at least in a preliminary, \textit{implicit} form). The first stage thus substantiates them as separate object-types, basic enough to play a role in daily life and be learnt. Ordinals are relevant for example wherever ordered operations are involved. Shifting the grasp (representation) of a song between its lines and its multi-line verses actually involves ordinal multiplication. Likewise for cardinality, a child may be able to match forks to plates when setting a table, without paying attention - representing - any particular order by which to do so. Moreover, she may understand that approaching the table from a different side doesn't change anything about the task. Such are feats that the cognitive system achieves. 

Given these acquired object-types, we can consider the interaction between them. Unlike the general case, in the \textit{restricted context} of the finite domain, their interaction presents a substantial regularity: We can move back and forth between an ordinal and its matching cardinal. And we can switch between ordinal arithmetic and cardinal arithmetic – these are coordinated accordingly. Falling right in line with the framework's abstract principles (omitted here), these object-types can be merged coherently into a higher-level (dimension-reduced) object-type -- that of numbers. This structure, which seamlessly integrates both as aspects, is precisely where the non-conscious, automated interplay that allows for "seeing" the commutativity of their multiplication resides. This is an instance of the value of the higher-level representation, towards which the system is driven. 

With this ontological structure clarified, we can consider how our developmental track revolves around it (beginning with the system's drive to amalgamate object-types when it can). For example, why would we (consciously) forget about ordinals and cardinals, while we do not forget of mammals nor winged animals upon learning of bats? How do teachers fit into this track? And more generally, how does this view square with the vast empirical research on the development of these abilities?

\section{Conclusion}

For mathematics as currently conceived, the ordinals-cardinals case study is a cautionary tale. The finite is the implicit context in which these mathematical objects happen to first be experienced and discovered or taught. The tight connection between them is in this sense a contingent, empirical regularity rather than an absolute mathematical necessity. Yet the cognitive system, by its general, generally-useful object-learning nature, comes to integrate it implicitly into our conception. Even logical codification is susceptible to the partiality of such explicit conceptions (as in coding the natural numbers through Peano axioms, which does not code the way they are used).

The more general lesson, however, I take to be more optimistic. Yes, mathematics is unique, not least because fundamental aspects of mathematical reasoning are unique. But mathematical reasoning as done by us humans, within the context of our full cognitive system, makes room for \textit{general} intelligence to enable and support it in various ways. Perceptual patterns, in particular, may run deep, integrated into mathematical reasoning pre-symbolically. This exploration holds the promise for advancing our understanding of humans’ understanding of mathematics, and through that, indeed, of (the realm of) mathematics itself (as I have set to demonstrate with a very particular example). But for that, we must approach it all in sufficiently abstract terms. After all, our ability to reason about this unique, newly-discovered realm, is a testimony to the extreme generality of our general intelligence. \citep{sloman2015welldesigned}

\subsubsection*{Acknowledgments}
The author thanks Amit Miller, Gilad Landau, and three anonymous referees for comments on this paper. And most of all, Carl Posy and Aaron Sloman, for supervising the PhD on which this paper is based.

\bibliography{iclr2021_conference}
\bibliographystyle{iclr2021_conference}

\end{document}